\documentclass[a4paper,twoside]{article}

\usepackage{epsfig}
\usepackage{subcaption}
\usepackage{calc}
\usepackage{amssymb}
\usepackage{amstext}
\usepackage{amsmath}
\usepackage{amsthm}
\usepackage{multicol}
\usepackage{pslatex}
\usepackage{apalike}
\usepackage{algorithm2e}
\usepackage{hyperref}
\usepackage{orcidlink}
\usepackage[bottom]{footmisc}
\usepackage{SCITEPRESS}     

\begin{document}

\title{Development and Application of a Sentinel-2 Satellite Imagery Dataset for Deep-Learning Driven Forest Wildfire Detection}
\author{
\authorname{
Valeria Martin\sup{1}\orcidlink{0009-0000-3668-5003}, 
K. Brent Venable\sup{2}\orcidlink{0000-0002-1092-9759}, 
and Derek Morgan\sup{3}\orcidlink{0000-0003-2321-3765}
}
\affiliation{\sup{1}
Department of Intelligent Systems and Robotics, University of West Florida, Pensacola, FL, USA. \email{vm58@students.uwf.edu}}
\affiliation{\sup{2}
Department of Intelligent Systems and Robotics, University of West Florida, and Florida Insitute for Human and Machine Cognition (IHMC), Pensacola, FL, USA. \email{bvenable@uwf.edu}}
\affiliation{\sup{3} Department of Earth and Environmental Sciences, University of West Florida, Pensacola, FL, USA. \email{jmorgan3@uwf.edu}}
}


\abstract{Forest loss due to natural events, such as wildfires, represents an increasing global challenge that demands advanced analytical methods for effective detection and mitigation. To this end, the integration of satellite imagery with deep learning (DL) methods has become essential. Nevertheless, this approach requires substantial amounts of labeled data to produce accurate results. In this study, we use bi-temporal Sentinel-2 satellite imagery sourced from Google Earth Engine (GEE) to build the California Wildfire GeoImaging Dataset (CWGID), a high-resolution labeled satellite imagery dataset with over 100,000 labeled \textit{before} and \textit{after} forest wildfire image pairs for wildfire detection through DL. Our methods include data acquisition from authoritative sources, data processing, and an initial dataset analysis using three pre-trained Convolutional Neural Network (CNN) architectures.
Our results show that the EF EfficientNet-B0 model achieves the highest accuracy of over 92\% in detecting forest wildfires. The CWGID and the methodology used to build it, prove to be a valuable resource for training and testing DL architectures for forest wildfire detection.}

\onecolumn \maketitle \normalsize \setcounter{footnote}{0} \vfill

\section{\uppercase{INTRODUCTION}}

\label{sec:introduction}

Forests are essential global ecosystems, offering habitats for countless species and playing a crucial role in sustaining environmental health by supporting biodiversity, regulating the climate, and producing oxygen. They also provide significant social and economic benefits, including energy production, job opportunities, and spaces for recreation. With increasing concerns over forest disturbance, safeguarding these ecosystems has become an urgent global priority \cite{IUCN2021}.
\\
\indent Forest monitoring capabilities have improved with the development of advanced satellite technology, including high-resolution imagery and its higher temporal resolution. For instance, the open source Sentinel-2 satellites constellation, operated by the European Space Agency (ESA), provide access to high-quality satellite images, with a revisit time of just 5 days, allowing for continuous surveillance of forested areas \cite{DRUSCH201225}. The development of Earth Observation (EO) systems has made remote sensing an efficient and cost-effective method for identifying and monitoring forest disturbance. \cite{Massey2023}.
\\
\indent Traditional approaches to forest disturbance detection using EO systems typically involve manually identifying feature classes and applying predefined algorithms or models, such as differential analysis and clustering techniques. These approaches require considerable domain knowledge and may fail to capture the data’s complexity \cite{rs14071552}.
\\
\indent Recent advances in deep learning (DL) algorithms, particularly in computer vision methods like Convolutional Neural Networks (CNNs) \cite{lecun} and Fully Convolutional Neural Networks (FCNs), such as the U-Net \cite{DBLP:RonnebergerFB15}, offer new ways to improve forest disturbance detection. These algorithms can identify complex patterns in large datasets, making them ideal for use with EO systems to monitor forests in near real-time. They also help in predicting and tracking different causes of forest disturbances, as well as their spread and impact \cite{eleo,al-dabbagh2023uni}.
\\
The success of DL algorithms in this context, however, depends on the availability of large, well-labeled datasets \cite{Alzubaidi2021ReviewOD}. Currently, the lack of high-quality labeled satellite imagery poses a challenge to the development of DL models for forest disturbance detection \cite{Adegun2023}. To address this issue, creating detailed datasets targeting specific forest disturbances, such as wildfires, is critical. This project focuses on building a satellite imagery dataset for detecting wildfires using DL techniques. 
\\
\indent Uncontrolled forest wildfires, often considered natural disasters, can cause widespread and long-lasting destruction \cite{berlinck2020good,hutto2008ecological,vilar2007analisis}. They can be responsible for significant losses of human lives and property damage each year. The frequency and intensity of these wildfires increased due to human activity and climate change, resulting in longer and more devastating wildfire seasons \cite{arteaga2020deep}. Hence, there is a vital need to improve prediction, mitigation, and forest wildfire disaster response. To do this, understanding the location, size, and frequency of wildfires in near real-time is essential. 
\\
\indent California has been particularly affected by forest wildfires over the years due to its vast forested areas and changing climate conditions, marked by prolonged droughts, higher temperatures, and shifting precipitation patterns \cite{li2021spatial}. This state also combines diverse vegetation types, from dense forests to shrublands, and varying topographies \cite{mcv2}. Additionally, California has historical forest wildfire perimeters from the Fire and Resource Assessment Program (FRAP) that outline wildfire-affected areas (see Fig. \ref{fig:polygon}). Therefore, this state offers the opportunity to study forest wildfires in a variety of ecological and historical contexts. 
\\
\indent Creating a satellite imagery dataset for California focused on wildfire perimeters can advance our understanding of forest wildfires. The specific characteristics of this geographical area make the dataset interesting and relevant for current and future applications as well as for investigating the transferability of models trained on it to other areas with similar features.
\\
\indent The following study describes the pipeline for building and testing the California Wildfire GeoImaging Dataset (CWGID), a high-resolution bi-temporal labeled satellite imagery dataset developed using Sentinel-2 imagery and Google Earth Engine (GEE) for DL-driven forest wildfire detection. 
\\
\indent This pipeline can serve as a baseline for building high-resolution labeled satellite imagery datasets for multiple forest disturbance causes. 

\section{RELATED WORK}
\indent 
In this section, we review some recent work closely related to what we present in this paper. 
\\
\indent Abujayyab et al. (2023), developed a CNN model to detect wildfires and trained it for 200 epochs using Sentinel-2 imagery from the Mediterranean region of Turkey. Their training dataset consisted of 159 fire images and 149 non-fire images. They achieved an accuracy of 92.5\% and a loss of 0.22 on the test set. In their research, the authors acknowledged the need for further performance metrics and larger datasets covering a broader geographical area to improve the generalization capability of their CNN model for wildfire detection \cite{abujayyab2023wildfire}. Our research expands upon this by building a much larger and more geographically diverse bi-temporal training dataset. Furthermore, while their model required 200 epochs to reach high accuracy results, our Early Fusion (EF) EfficientNet-B0 approach achieves a higher accuracy and a lower loss with 13 training epochs. In addition, our study incorporates additional performance metrics beyond accuracy and loss, such as precision and recall.
\\
\indent Elizaroshan et al. (2024), developed a web-based wildfire detection system using the DenseNet201 architecture, trained on a Kaggle satellite imagery dataset, achieving an accuracy of 92.39\% for wildfire detection. Their approach focused on uni-temporal satellite imagery and employed data augmentation to improve model generalization. However, they acknowledged limitations in dataset diversity, which may reduce the model’s ability to generalize accurately across diverse environmental settings. In contrast, our study leverages a bi-temporal dataset using Sentinel-2 satellite imagery, incorporating both pre- and post-wildfire images to capture the changes in forest conditions over time. Our CNN-based model, particularly the EF EfficientNet-B0, trained on the CWGID, outperformed their approach, achieving higher accuracy on the test set. Moreover, the CWGID includes a large and diverse set of wildfire examples.
\\
\indent Zhang et al. (2024), study the integration of multi-source satellite data and deep learning techniques for large-scale wildfire detection, leveraging bi-temporal imagery from Sentinel-2, Sentinel-1, and ALOS-2 PALSAR-2. The study shows the advantages of using bi-temporal data over uni-temporal data for wildfire detection, particularly using Sentinel-2.  Furthermore, they explain that while  SAR data from Sentinel-1 and ALOS-2 were valuable, imagery from Sentinel-2 was more effective for capturing burned areas and achieved higher performance results, due to its high spatial resolution (10m) and spectral coverage. \cite{zhang2024assessing}. This aligns with our methodology, which leverages a large and diverse bi-temporal Sentinel-2 dataset to capture the changes between pre- and post-wildfire events.
\\
\indent Al-Dabbagh and Illyas (2023), developed a uni-temporal Sentinel-2 post-wildfire imagery dataset for DL applications using imagery from the 2021 wildfire season in Turkey. Their dataset comprises imagery from five tiles captured in September 2021 and sourced from different areas in Turkey. They download the images individually from the United States Geological Survey (USGS) platform and they label them manually using ArcGIS. Furthermore, they segment their tiles into 128*128px images, resulting in 21,690 images multi-band GeoTIFF images\cite{al-dabbagh2023uni}. The CWGID is also developed using Sentinel-2 imagery but a bi-temporal approach of pre- and post-wildfire imagery is used to build it. The CWGID allows the use and comparison of uni-temporal and bi-temporal approaches with DL. This study shows that using \textit{before} and \textit{after} wildfire images as coupled inputs improves performance with respect to the uni-temporal case (see Section \ref{sec:result}). Moreover, the CWGID has data from multiple wildfire seasons (from 2016 to 2021) spanning various forested areas in California. This enhances its utility for DL models by offering more examples of wildfire dynamics over time and in different vegetated areas. Additionally, the CWGID is created by downloading and labeling images programmatically using GEE and existing curated data, which constitutes a time-effective approach. Finally, the CWGID has many more examples of wildfires, with more than $100,000$ 256*256px GeoTIFF \textit{before} and \textit{after} wildfire image pairs. 
\\
\indent Xiang et al. (2023), developed a multi-temporal Sentinel-2 RGB imagery dataset to detect forest change in the Hunan province using DL models. Similar to the CWGID, this dataset is built using GEE, and the ground-truth labels are programmatically rasterized using existing data and ArcMap. They source 20 image tiles and crop them into 1437 256*256px image pairs.  Moreover, they test their data using well-known FCN architectures, such as UNet++ achieving an average precision of 75\% \cite{Hunan}. While their methodology is similar to the one we use for building the CWGID, our approach using CNNs outperforms their results. Specifically, the Early Fusion EfficientNet-B0 model in our study achieves a precision of 92.5\%, and the Siamese EfficientNet reaches 87.5\%. Also, the CWGID consists of many more 256*256px image pairs for DL training. 
\\
\indent Generally, multiple DL architectures have been developed for wildfire detection and trained using examples from Sentinel-2 satellite imagery \cite{brand2021semantic,knopp2020deep,zhang2021deep}. However, 
the datasets used to train these models often fail to capture the complex dynamics of wildfires across varied terrains and vegetation types. This can result in models that are not fully optimized for the diverse conditions wildfires present. To overcome these limitations, the CWGID focuses on compiling a larger, more comprehensive dataset that encompasses bi-temporal imagery. 

\indent In addition to the research mentioned above, wildfire detection has also been tackled using FCNs, such as UNet, U-Net++, and DeepLabV3+ (\cite{pinto2021practical,SEYDI2022108999,sun2022analyzing}). In this paper, however, we focus on CNN models that provide prediction at the image level rather than at the pixel level.  
We also note that other wildfire datasets for DL-driven wildfire detection have been recently built using non-satellite imagery, such as aerial photography, but these studies are beyond the scope of this paper \cite{alice2023automated,davis,deepfire,xie2016deep3d}. 
\\
\section{\uppercase{METHODS}}
\begin{figure*}[!h]
  \centering
   \includegraphics[width=15cm]{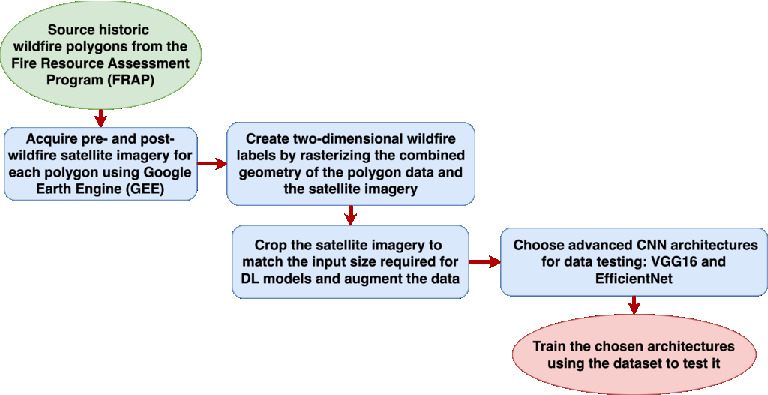}
  \caption{Flowchart of the proposed methodology. The diagram illustrates the sequential steps of the workflow followed in this study.}
  \label{fig:flow}
  \end{figure*}

In this section, we present a complete workflow of the end-to-end process of creating and using our forest wildfire imagery dataset, from data acquisition to the application of DL for change detection. The flowchart depicted in Fig. \ref{fig: flow} outlines the step-by-step process of the proposed methodology.  
\subsection{Data Creation}
Building labeled satellite imagery datasets for DL-driven forest disturbance detection, such as the CWGID, involves a series of steps.
\subsubsection{Gathering and Refining Wildfire Data}
\begin{figure*}[!h]
  \centering
   \includegraphics[width=15cm]{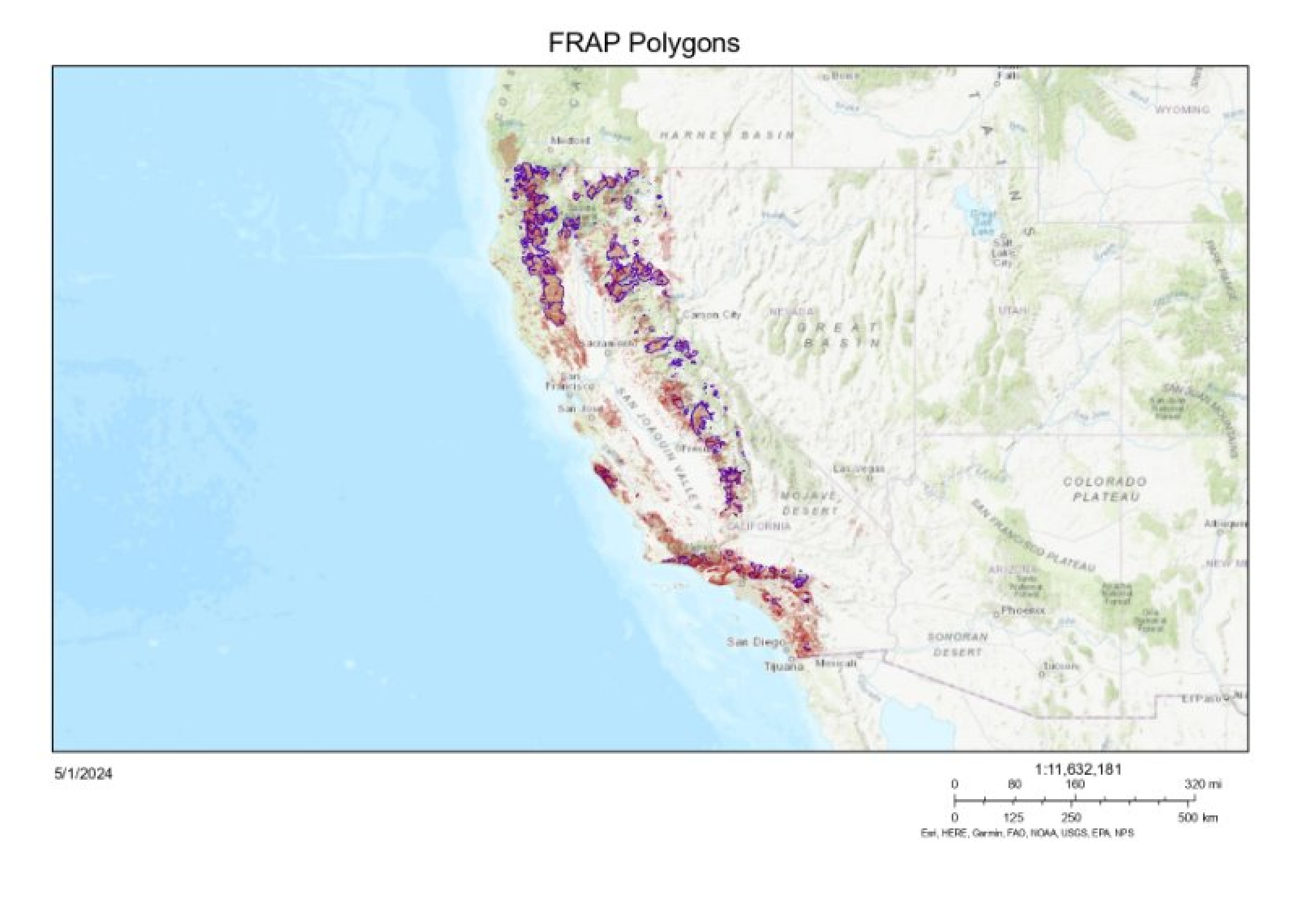}
  \caption{Representation of the Polygon Data from the FRAP. Polygons in purple represent wildfires in forested areas, used for the CWGID.}
  \label{fig:polygon}
  \end{figure*}

The initial step in the process of building the CWGID is to gather curated wildfire polygons from California that are sourced from FRAP. Next, this shapefile data is filtered in ArcGIS Pro to focus on wildfires, excluding other types of fires, such as prescribed burns. Further refinement is done by clipping the wildfire polygon layer to forested areas using the Original Proclaimed National Forest Land layer from the United States Forest Service  (see Fig. \ref{fig:polygon}). Then, an Excel file containing the attribute table data and the wildfire centroids is exported from ArcGIS Pro for further processing.

\subsubsection{Date Filtering and Satellite Selection}

The Excel file containing wildfire centroids undergoes a filtering process in Python to select data from the years 2016 to 2021. This period is chosen to align with the operational phase of the Sentinel-2 satellites. 
\\
\indent The Sentinel-2 satellite imagery is chosen to create the forest wildfire imagery dataset due to its unrestricted access, its high frequency revisit times of 5 days (with the combined constellation) and its high spatial resolution of 10 meters \cite{DRUSCH201225}, which is significant when compared with other publicly available satellite imagery such as Landsat (30 m) and MODIS (250–500 m). 
\\
\indent Further scripting is needed to adjust the wildfire dates to fall within the green-up period, avoiding the winter and fall seasons where snow cover can interfere with the identification of burnt areas. 

\subsubsection{Image Downloads with Google Earth Engine}
GEE is a cloud-based platform that provides high-performance computational tools for analyzing environmental data on a global scale \cite{gorelick2017google}. GEE enables researchers to monitor and detect changes on the Earth’s surface by offering access to an extensive archive of satellite imagery and geospatial datasets. Its Python API streamlines the process of automating satellite imagery downloads, making it an invaluable tool for environmental monitoring and research efforts.
\\
\indent Users can query the GEE database for specific satellite images based on different parameters, such as geographic location, time range, and cloud coverage. This capability is particularly useful for projects that require the effective collection of large datasets over specific areas of interest.
\\
\indent In the context of the CWGID, GEE's Python API is utilized to automate the download of \textit{before} and \textit{after} wildfire Sentinel-2 satellite images. To do this, adjusting the wildfire perimeter data to comply with GEE's guidelines is needed. 
\\
\indent Firstly, for querying purposes, the wildfire centroid coordinates are transformed from the North American Datum 1983 (NAD83) to the World Geodetic System 1984 (WGS84). Then, a 15-day range for both pre-wildfire and post-wildfire dates is generated and formatted to meet GEE's requirements. Furthermore, a squared region of interest (ROI), featuring a side length of 15 miles, is defined and centered around the forest wildfire centroid coordinates. This step is important to define the standard size of the satellite imagery to be downloaded. 
\\
\indent Also, the satellite imagery bands to be used and downloaded are defined and stored. In this study, the bands used are B4, B3, and B2 to build three-channel GeoTIFF RGB imagery. Finally, the queried tiles are filtered and those with less than 10 \% cloud coverage are downloaded. 
\subsubsection{Ground Truth Mask Creation}
\begin{figure*}[!h]
  \centering
   \includegraphics[width=15cm]{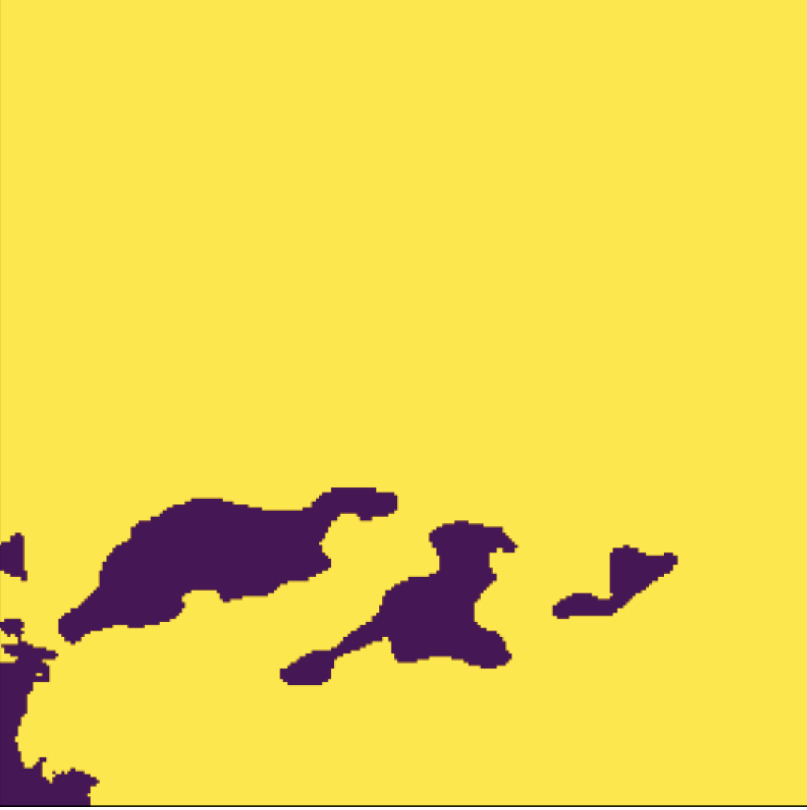}
  \caption{Example of a ground truth mask from the CWGID. The mask highlights wildfire-affected areas in purple and unaffected areas in yellow.}
  \label{fig:mask}
 \end{figure*}
Accurate forest wildfire detection and land cover classification heavily rely on the use of accurate ground truth masks \cite{8113128}. In this project, these masks are generated to serve as the two-dimensional labels needed to train and validate DL models. Particularly, these two-dimensional labels are used to distinguish between wildfire-affected and wildfire-unaffected areas. Hence, these masks are binary images where pixels with a value of 1 represent wildfire damage and pixels with a value of 0 represent the absence of it.  Ground-truth masks allow DL models to learn the characteristics of forest wildfires, such as their shape, size, and texture, which are crucial for accurate detection and assessment.
\\
\indent In this study, ground truth masks are created by overlaying the downloaded satellite imagery with the FRAP wildfire perimeters. This process involves intersecting the geometry of the satellite imagery with the geometry of the forest wildfire polygons and rasterizing the combined geometry. Fig. \ref{fig:mask} displays an example of a ground truth mask from the CWGID. The creation of masks is important for accurate and reliable detection of wildfires.

\subsubsection{Imagery Segmentation and Data Augmentation}
\begin{figure*}[h]
  \centering
    \includegraphics[width=15cm]{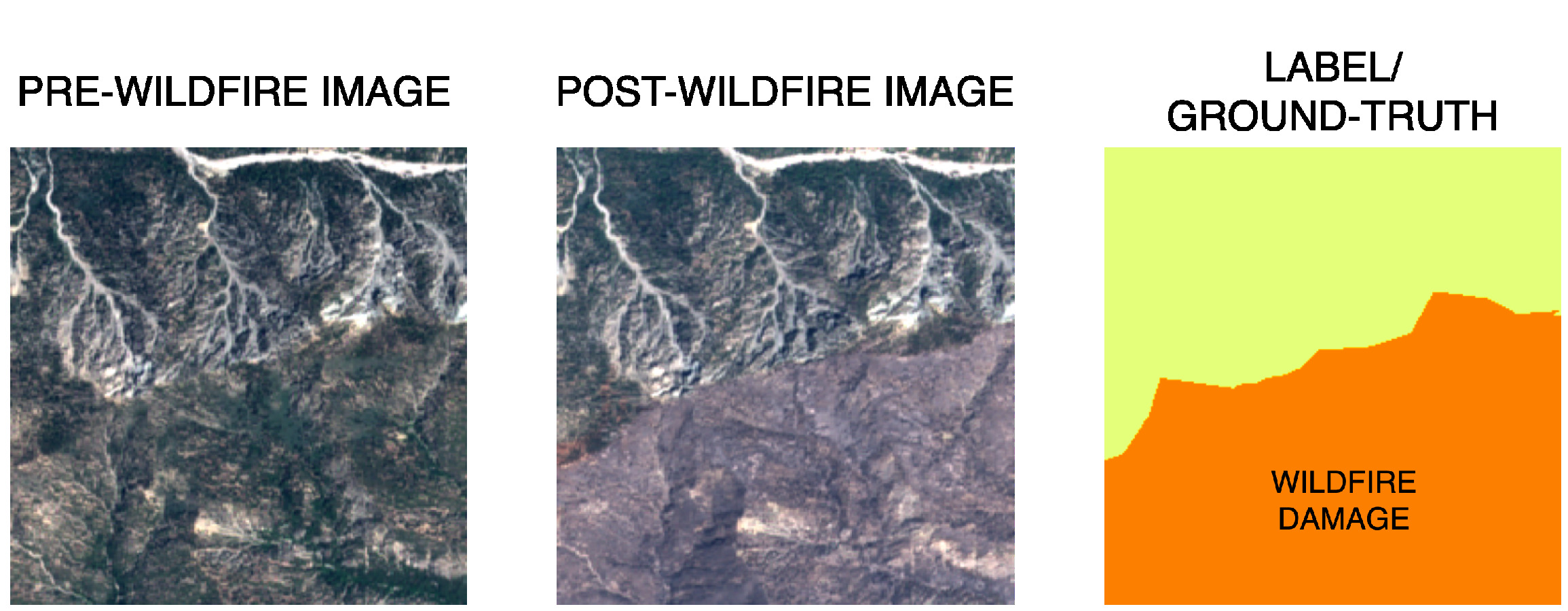}
  \caption{Example of 256*256px pre- and post-wildfire RGB tiles and their corresponding ground-truth mask.}
  \label{fig:crop}
 \end{figure*}
The satellite imagery and the ground-truth masks are segmented into smaller 256*256px RGB GeoTIFF tiles (see Fig, \ref{fig:crop}). This step is essential because satellite imagery is usually resized and downscaled to fit DL architectures, which can lead to the loss of critical information, such as small but significant features indicative of early-stage wildfires or detailed vegetation health. Furthermore, smaller image tiles are more efficiently processed by DL models because they reduce computational costs and improve training times \cite{hu2015transferring,marmanis2016deep}. This efficiency is vital for analyzing large datasets, common in satellite imagery analysis. Moreover, by focusing on smaller areas, models can better learn local patterns and features relevant to wildfire detection. Smaller tiles can ensure that DL models are not overwhelmed with the significant and varied information present in larger satellite imagery. 
\\
\indent It is important to mention the segmentation carried out in this project uses multiple Python libraries to ensure the integrity of the RGB data and to maintain georeferencing information.
\\
\indent Finally, the resulting imagery with positive instances of wildfire damage was augmented by applying data augmentation techniques. This included 3 rotations and horizontal and vertical flips. This process is particularly important as it introduces variations in the training data, can reduce overfitting, and reduces class imbalance, ensuring the model does not become biased towards the more common, unaffected landscapes \cite{perez2017effectiveness,shorten19}.
\\
\indent CWGID, the dataset obtained by following the workflow described above, comprises 106,317 pairs of labeled bi-temporal RGB GeoTIFF image tiles, with 29,082 positive instances (wildfire damage). 
The full dataset can be accessed via the following link\footnote{Dataset available at: \url{https://drive.google.com/drive/folders/13TWt7r-TPBOEJsiPvYqtDc-0JTyr4c6f?usp=sharing}}.

\subsection{Evaluating DL Architectures using CWGID}
\label{arch-sec}
\begin{figure*}[h]
  \centering
   \includegraphics[width=15cm]{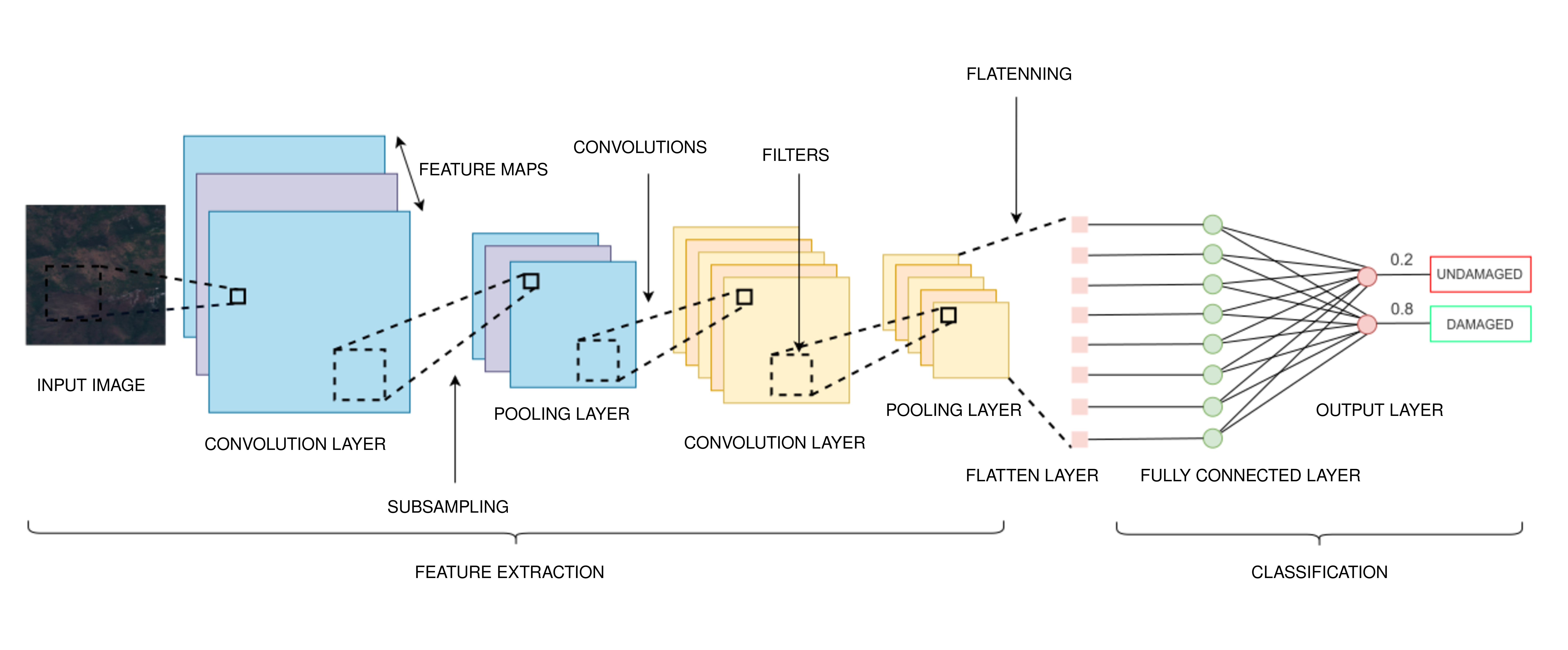}
  \caption{Representation of a CNN with an input image, two convolutional layers, two pooling layers, one fully connected layer, and the output layer. The output has 2 different classes: damaged and undamaged, for which we show an example of classification scores.}
  \label{fig:2}
 \end{figure*}

Automating the processing and classification of satellite images is crucial for efficiently detecting and mapping changes in forested areas. In recent years, Machine Learning (ML), and especially DL have proved to be useful methods for tackling these challenges \cite{8113128}.
\\
\indent DL is a subset of machine learning that uses neural networks with many layers to automatically learn and model complex patterns from large amounts of data. CNNs are a specialized DL architecture tailored for tasks like image processing and classification. Fig. \ref{fig:2}, for example,  depicts a CNN that processes an input image through two convolutional layers and two pooling layers, followed by a fully connected layer, culminating in an output layer. The output layer categorizes the image into two classes, wildfire-damaged and undamaged forests.

\indent In what follows we describe the implementation,  training and testing of  several CNN architectures, including VGG16 and EfficientNet, for wildfire detection using CWGID.

We note that all implementations were developed  with Python's TensorFlow Keras API and that the models were run locally on a MacBook Pro equipped with a 2.4 GHz Intel Core i9 processor and 16 GB of 2400 MHz DDR4 RAM.
 
\subsubsection{VGG16}
VGG16 \cite{Simonyan15} is a deep CNN model specifically designed for tasks involving image processing and classification. The architecture is structured with 13 convolutional layers that extract features from the input images, followed by 5 pooling layers that downsample the feature maps. Finally, the last 3 layers of the network are fully connected layers that perform the final classification. The convolutional layers use a series of 3x3 filters and the pooling layers operate on 2x2 pixel windows, preserving the largest element of the feature map. The first two fully connected layers have 4096 channels each, and the third performs the classification and has as many channels as there are classes in the dataset. This combination of layers allows VGG16 to effectively learn complex visual patterns in large datasets. This architecture uses the ReLU activation function \cite{goodfellow2016deep}. 
\\
\indent We adapted the VGG16 to train it and test it on the CWGID and to detect positive and negative instances of forest wildfires. The VGG16 architecture only trains on the \textit{after} wildfire 3-channel RGB imagery, which contains examples of both positive and negative instances. 
\\
\indent To apply this CNN architecture to GeoTIFF satellite images, which often include geographic metadata not supported by most deep learning libraries, a custom function to feed the data to the CNN architecture is built (see Algorithm 1). This function is important for preprocessing and feeding the data into VGG16 for both training and evaluation purposes.
\begin{algorithm}[!h] 
 \label{alg1}
 \caption{Custom Data Generator for VGG16 with GeoTIFF files.}
 \KwData{List of paths to GeoTIFF files (file\_paths), Batch size (batch\_size)}
 \KwResult{Generator yielding batches of image data and labels}
 initialization\;
 \While{True}{
  Shuffle file\_paths\;
  \For{each batch in file\_paths}{
   Initialize empty lists for images and labels\;
   \For{each file in the batch}{
    Read image from file\;
    Move the axis to convert channels from the first to last dimension\;
    Assign label based on the presence of 'Damaged' in file path\;
    Append image and label to their respective lists\;
   }
   Yield a tuple of image and label arrays\;
  }
 }
\end{algorithm}
The data labeling and its use are established by specifying the base paths to the training and testing directories for both damaged and undamaged classes. The custom image generator reads previously defined batches of GeoTIFF files, shuffles them, and processes them into a three-dimensional array compatible with VGG16. Since the VGG16 model simply needs to discern the presence or the absence of wildfire damage in the image, the ground-truth masks are not directly used. They are only used to define the imagery file paths. Thus, the labeling of this model is based on the presence of \textit{Damaged} (value of 1) in the imagery file paths. All the other files are labeled as \textit{Undamaged} (value of 0), simplifying the binary classification task. 
\\
\indent The VGG16 model is pre-loaded with weights from the ImageNet dataset and used to detect wildfire-affected regions. To maintain the integrity of the features learned during pre-training, the convolutional base is frozen, shifting the training focus to the newly added layers. The two-dimensional feature maps produced by the network are then transformed into a one-dimensional vector using a \textit{Flatten} operation. Furthermore, a fully connected \textit{dense} layer with 1024 neurons and ReLU activation is added. Finally, the network is configured to end with a single-neuron \textit{dense} layer that uses the \textit{sigmoid} activation function. This layer outputs a probability score indicating the likelihood of wildfire damage in the input image. 
\\
\indent Additionally, the Adaptive Moment Estimation (Adam) optimizer \cite{kingma2014adam} is employed, chosen for its adaptive learning rate capabilities, which are beneficial in satellite image classification tasks. Since satellite images can exhibit significant variability in terrain and features, Adam's ability to dynamically adjust the learning rate enhances the model's performance over diverse datasets. The model is also compiled with binary cross-entropy loss \cite{goodfellow2016deep}, a loss function suited for binary classification tasks, like distinguishing between forest wildfire-damaged and undamaged areas. Furthermore, accuracy, precision, and recall are chosen as performance metrics, providing an evaluation of the model’s ability to correctly classify wildfire-affected regions after training with the CWGID. This combination of optimizer, loss function, and metrics ensures the model is well-equipped to handle the challenges associated with detecting wildfire damage in different geographic settings.
\\
\indent The VGG16 model is trained and validated on 95594 three-channel RGB images (45\% of the CWGID, divided as follows: 80\% for training and 10\% for validation) and is tested on 10621 images (5\% of the CWGID). Training took 1953 minutes for 10 epochs. We provide the results of this model in Section \ref{sec:result}.

\subsubsection{EfficientNet}
\begin{algorithm}[!h]
\label{alg2}
 \caption{Custom Data Generator for Multi-Channel Images.}
 \KwData{List of image file paths (file\_paths), Image labels (labels), Batch size (batch\_size), Image dimensions (dim), Number of channels (n\_channels), Shuffle indicator (shuffle)}
 \KwResult{Generator yielding batches of six-channel image data and labels}
 
 \SetKwProg{Fn}{Function}{:}{}
 \SetKwFunction{FMain}{SixChannelGenerator}
 \SetKwFunction{FLength}{\_\_len\_\_}
 \SetKwFunction{FGetItem}{\_\_getitem\_\_}
 \SetKwFunction{FEpochEnd}{on\_epoch\_end}
 
 \Fn{\FMain{file\_paths, labels, batch\_size, dim, n\_channels, shuffle}}{
  Initialize file\_paths, labels, batch\_size, dim, n\_channels, shuffle\;
  \FEpochEnd{}\;
 }
 
 \Fn{\FLength{}}{
  \KwRet $\lceil$len(file\_paths) / batch\_size$\rceil$\;
 }
 \Fn{\FGetItem{index}}{
  batch\_paths $\leftarrow$ file\_paths[index $\cdot$ batch\_size : (index + 1) $\cdot$ batch\_size]\;
  batch\_labels $\leftarrow$ labels[index $\cdot$ batch\_size : (index + 1) $\cdot$ batch\_size]\;
  batch\_x $\leftarrow$ empty array of shape (len(batch\_paths), dim[0], dim[1], n\_channels)\;
  batch\_y $\leftarrow$ array of batch\_labels\;
  
  \For{i $\leftarrow$ 0 \KwTo len(batch\_paths)}{
   Read image from batch\_paths[i]\;
   Select first n\_channels\;
   Move axis to convert from channels\_first to channels\_last format\;
   Normalize image and assign to batch\_x[i,]\;
  }
  
  \KwRet batch\_x, batch\_y\;
 }
 
 \Fn{\FEpochEnd{}}{
  \If{shuffle}{
   Combine file\_paths and labels into a single list\;
   Shuffle the combined list\;
   Unpack shuffled list back into file\_paths and labels\;
  }
 }
\end{algorithm}

\begin{figure*}[!h]
  \centering
   \includegraphics[width=15cm]{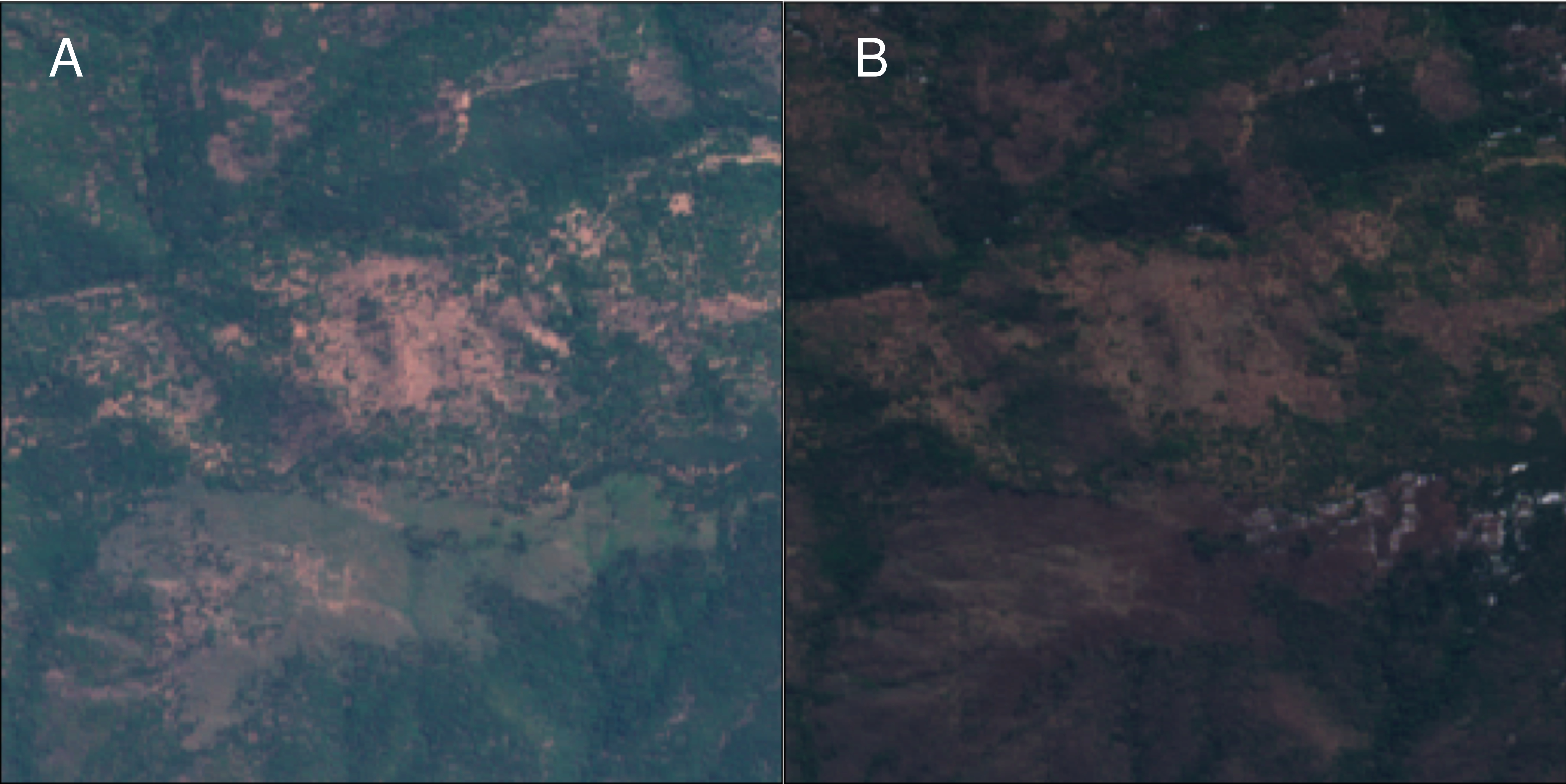}
  \caption{Representation of a 6 Channel RGB GeoTIFF input. A: Representation of a 3-channel RGB GeoTIFF forested area \textit{before} a wildfire B: Visual example of a 3-channel RGB GeoTIFF forested area \textit{after} a wildfire.}
  \label{fig:3}
 \end{figure*}
EfficientNet \cite{tan2019efficientnet} is a CNN architecture that uniformly scales network width, depth, and resolution with a fixed set of scaling coefficients. 
\\
\indent EfficientNet's approach allows the optimization of both performance and computational resources, achieving state-of-the-art accuracy on image classification tasks with high efficiency.  
\\
\indent EfficientNet's architecture starts with the base model, EfficientNet-B0, which is specifically designed to identify the most efficient baseline network configuration. Subsequent versions of the model, such as B1 through B7, represent scaled variations of B0, offering a range of options tailored for different computational resources and performance requirements.
\\
\indent The EfficientNet-B0 architecture is selected for this project. This architecture consists of a base convolutional layer that processes the input image by extracting initial features from the raw input. Then, at the network's core, there is a series of seven Mobile Inverted Bottleneck Convolution (MBConv) blocks. Each block consists of a bottleneck architecture that includes: a 1x1 convolution for channel size expansion, a depthwise 3x3 convolution for spatial feature extraction, and another 1x1 convolution to project the features to a lower-dimensional space. After each MBConv block, there is a global average pooling layer that reduces the feature maps into a single mean scalar. Similar to the VGG16 architecture, a \textit{Flatten} operation is applied to transform the two-dimensional feature maps into a one-dimensional vector. Next, a fully connected \textit{dense} layer is added. Finally, the network becomes a single-neuron \textit{dense} layer with a \textit{sigmoid} activation function that serves as the classifier part of the network.  
\\
\\
\indent For this project, Efficient-B0 is adapted, trained, and tested using \textit{before} and \textit{after} wildfire RGB GeoTIFF imagery pairs from the CWGID. To do this, two methodologies are employed. 
\\
\\
\indent The first method employed is called Early Fusion (EF), where image pairs are combined into six-channel GeoTIFF files to form a single input for the model (see Fig. \ref{fig:3}). A custom function to feed the data to the architecture is built to allow EfficientNet-B0 to process these six-channel GeoTIFFs (see Algorithm 2). Labeling for this approach is done as for VGG16, by setting base paths for the training and testing directories of both the 'damaged' and 'undamaged' classes. The image file paths are checked for the term 'Damaged' to assign the appropriate label.
\\
\indent The images are first processed by a convolutional layer that performs the initial convolution operation. The base EfficientNet-B0 model is then loaded, excluding the top classification layers, as the pre-trained weights were developed using 3-channel images. To address potential overfitting, a 30\% Dropout layer is added. A dense layer with 1,024 neurons and ReLU activation is also included, and the network outputs a probability score indicating whether the input image is 'damaged' or 'undamaged.'
\\
\indent Class weights are calculated to handle the imbalance between 'damaged' and 'undamaged' images, ensuring balanced learning. Several callbacks are employed, including early stopping, model checkpointing, and learning rate reduction when validation loss plateaus, optimizing the training process. The model is compiled with the Adam optimizer and binary cross-entropy loss, and accuracy, precision, and recall are chosen as performance metrics, similar to the setup for VGG16.
\\
\indent For testing purposes, this EF architecture is trained on 23833 \textit{before} \textit{after} wildfire image pairs (22.5\% of the CWGID, divided as follows: 80\% for training and 10\% for validation) and is tested on 2716 \textit{before} \textit{after} image pairs (2.5\% of the CWGID). Training took 933 minutes for 13 epochs. To see the results of this model refer to Section \ref{sec:result} of this document.
\\
\begin{figure*}[h]
  \centering
   \includegraphics[width=15cm]{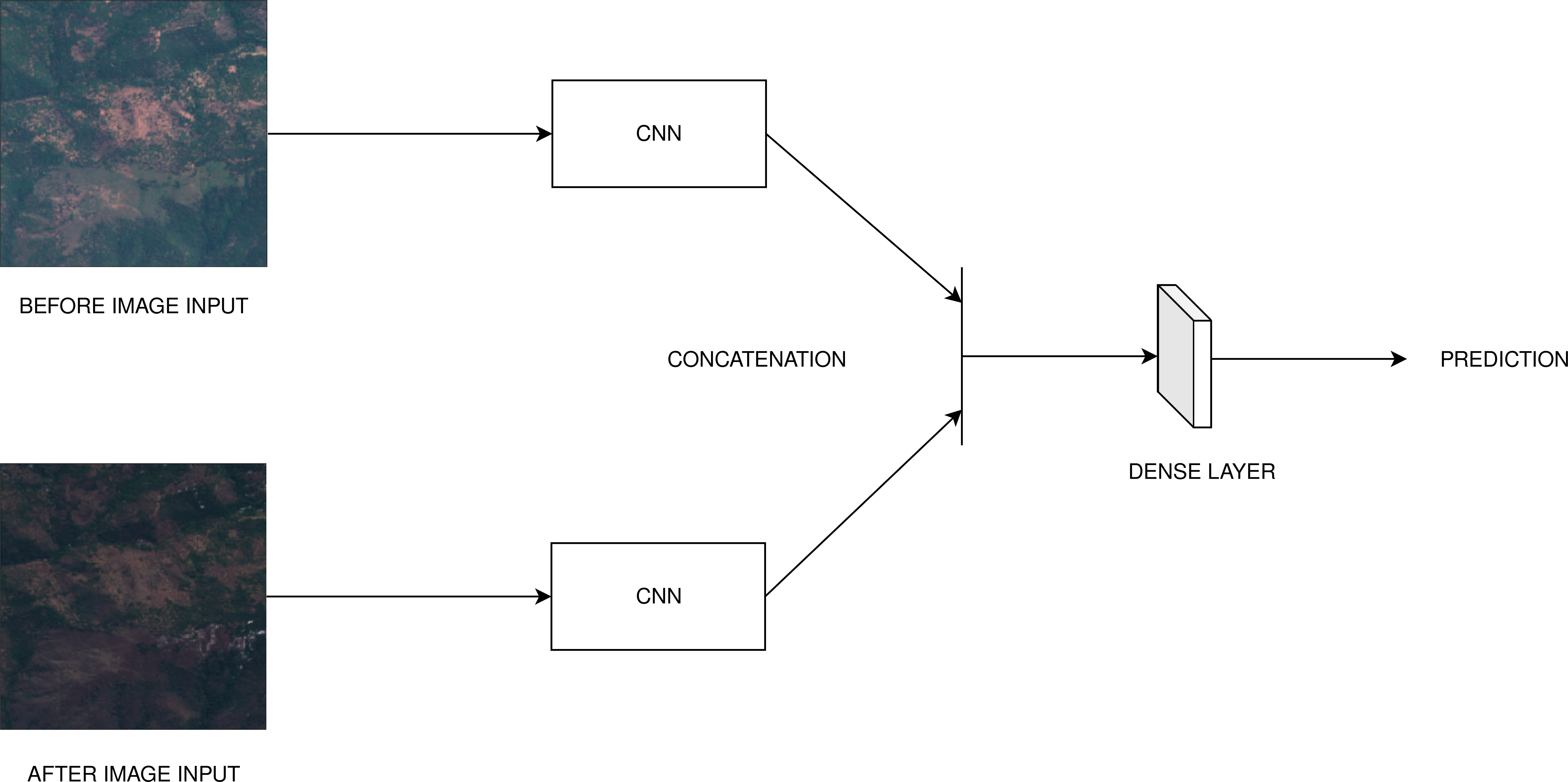}
  \caption{Representation of a Siamese network. The network has two inputs, two parallel CNN architectures, a concatenation, and a dense layer that produces the probability outputs.}
  \label{fig:4}
 \end{figure*}
\\
\indent In the second approach, a Siamese network \cite{bromley} is defined using two input branches, one for \textit{before} images and one for \textit{after} images (see Fig. \ref{fig:4}). A custom data generator for this structure is also defined (see Algorithm 3). As the models above, the data labeling and its use are defined on the file paths. The presence of ’Damaged’ in the imagery file paths corresponds to a value of 1, and the absence of it corresponds to a value of 0. 
\\
\indent This architecture uses the same 6-channel GeoTIFF files but the data generator feeds the first three channels of the file to the first branch (corresponding to the \textit{before} wildfire image) and the last three channels of the file to the second branch (corresponding to the \textit{after} wildfire image) (see Fig. \ref{fig:3}).

\begin{algorithm}[!h]
\caption{Siamese Network Data Generator.}
\KwData{List of image paths (image\_paths), Corresponding labels (labels), Batch size (batch\_size)}
\KwResult{Generator yielding pairs of 'before' and 'after' images with labels}
\SetKwProg{Fn}{Function}{:}{}
\SetKwFunction{FMain}{SiameseGenerator}
\SetKwFunction{FLength}{\_\_len\_\_}
\SetKwFunction{FGetItem}{\_\_getitem\_\_}
\SetKwFunction{FEpochEnd}{on\_epoch\_end}

\Fn{\FMain{image\_paths, labels, batch\_size, target\_size}}{
    Initialize image\_paths, labels, batch\_size\;
    \FEpochEnd{}\; 
}
\Fn{\FLength{}}{
    \KwRet $\lfloor$len(image\_paths) / batch\_size$\rfloor$\;
}
\Fn{\FGetItem{index}}{
    batch\_paths $\leftarrow$ image\_paths[index $\cdot$ batch\_size : (index + 1) $\cdot$ batch\_size]\;
    batch\_labels $\leftarrow$ array of labels[index $\cdot$ batch\_size : (index + 1) $\cdot$ batch\_size]\;
    Initialize batch\_before and batch\_after arrays with zeros\;
    \For{i $\leftarrow$ 0 \KwTo len(batch\_paths)}{
        Read image from batch\_paths[i]\;
        First 3 channels $\rightarrow$ batch\_before[i], normalize to [0, 1]\;
        Next 3 channels $\rightarrow$ batch\_after[i], normalize to [0, 1]\;
    }
    \KwRet [batch\_before, batch\_after], batch\_labels\;
}
\Fn{\FEpochEnd{}}{
    Shuffle image\_paths and labels while maintaining correspondence\;
}
\end{algorithm}

EfficientNet-B0 is used again as the feature extractor, loaded with pre-trained ImageNet weights and set to non-trainable to keep the learned features fixed. Outputs of the two EfficientNet-B0 branches are concatenated and followed by dense layers with ReLU activation to combine the features and make predictions. The final layer uses a sigmoid activation function to output a probability, indicating whether the input pair shows damage.
\begin{table*}[h] 
\caption{Performance of the DL architectures.}
\label{tab:1}
     \centering
     \begin{tabular}{|c|c|c|c|} \hline 
                      & VGG16 & 6-channel input EfficientNet-B0 & Siamese EfficientNet-B0  \\ \hline 
Loss                 & 1.220  & 0.178                           & 0.413 \\ \hline 
Accuracy             & 0.832  & 0.926                           & 0.816 \\ \hline 
Precision            & 0.749  & 0.925                           & 0.875 \\ \hline 
Recall               & 0.588  & 0.819                           & 0.442 \\ \hline 
Time (minutes)       & 1953   & 933                             & 1007 \\ \hline
     \end{tabular}
\end{table*}
 \\
\indent Training is performed using early stopping, model checkpointing, and learning rate reduction as callbacks. Accuracy, precision, and recall are included as performance metrics. Class weights are not calculated in this Siamese model because the accuracy dropped with this model when doing so. 
\\
\indent For testing purposes, the Siamese architecture is trained on 23833 \textit{before} \textit{after} wildfire image pairs (22.5\% of the CWGID, divided as follows: 80\% for training and 10\% for validation) and was tested on 2716 \textit{before} \textit{after} image pairs (2.5\% of the CWGID). Training took 1007 minutes for 17 epochs. To see the results of this model refer to Section \ref{sec:result}.

\section{\uppercase{RESULTS}} \label{sec:result}

We recall that CWGID, in its entirety, consists of  106,317 pairs of labeled bi-temporal RGB GeoTIFF image tiles, of which  29,082 are positive instances featuring wildfire damage. 
\\
\indent Table \ref{tab:1} provides a comparison of the performance of the three CNN architectures described in Section \ref{arch-sec} and trained on the CWGID. Performance metrics such as loss, accuracy, precision, recall, and training time are evaluated for each model. Additionally, the table indicates the percentage of the CWGID dataset used for training each architecture.
\\
\indent All models are trained with subsets of the CWGID dataset, with the 6-channel input EF EfficientNet-B0 and the Siamese EfficientNet-B0 models using 25\% of the data. In contrast, the VGG16 model was trained on a larger subset, using 50\% of the data. 
\\
\indent VGG16, despite processing a larger dataset, had the highest loss (1.220) but maintained an accuracy of 83.2\%. Its precision (74.9\%) was the lowest among the three models. However, its recall of 58.8\% was higher than the Siamese EfficientNet's. This architecture had the longest relative training time.
\\
\indent The 6-channel input EfficientNet-B0 achieved the lowest loss (0.178) and highest accuracy (92.6\%), making it the most effective model in terms of overall performance. Its precision was also high at 92.5\%, and its recall was 81.9\%. 
\\
\indent The Siamese EfficientNet-B0 reported a loss of 0.413 and an accuracy of 81.6\%, with a precision of 87.5\%. However, its recall of 44.2\% was significantly lower than the other two models.
\\
\indent The EfficientNet-based models show a relatively faster training time when considering the epochs and the amount of data used. 
\\
\indent We also note that we investigated training VGG16 with Six-Channel inputs and that it proved to be unfeasible due to its significant computational cost.

\section{\uppercase{DISCUSSION}}

Our structured approach to generating a large, labeled dataset,
allowed us to use existing resources, reducing the need for manual labeling. This resulted in one of the largest datasets available for 
wildfire detection, comprising 106,317 pairs of labeled bi-temporal RGB GeoTIFF image tiles.

We now turn our attention to the performance of the DL architectures on the CWGID.
From Table \ref{tab:1} we can see that VGG16 demonstrates a higher recall than the Siamese EfficientNet-B0, indicating a better balance between identifying true positives and reducing false negatives. However, the recall remains relatively low for both models, suggesting room for improvement in capturing all wildfire-affected areas. This performance also comes with an increase in training time, highlighting that while VGG16 is more effective in identifying wildfire-affected regions than the Siamese architecture, it requires significantly more resources to do so.
\\
\indent The Siamese EfficientNet-B0, shows a higher precision in its predictions than VGG16, but exhibits a notably lower recall when compared to the other architectures, suggesting that it tends to miss a significant portion of true positives. This could imply that the model struggles to generalize over more subtle or challenging wildfire cases.
\\
\indent On the other hand, the 6-channel EfficientNet-B0 achieves high overall performance, indicating its ability to correctly identify damaged areas while maintaining fewer missed detections compared to the other architectures. The recall of the EfficientNet-B0 Early Fusion (EF) model is notably higher than the other architectures, indicating that it is better at identifying a larger portion of wildfire-affected areas. Its more efficient training time, combined with its handling of multi-channel inputs, shows that for this project, it is the most suitable for training on the CWGID, where both accuracy and computational efficiency are key.
\\
\indent Hence, the CWGID enabled CNN models to learn and identify the signs of wildfire damage with high accuracy, specifically with the EF EfficientNet-B0 model. Additionally, it is notable that we obtain 92\% accuracy with training on less than 25\% of the CWGID. The dataset's bi-temporal collection of \textit{before} and \textit{after} wildfire satellite imagery also offers the possibility to use different and accurate approaches for detecting forest wildfires with satellite imagery.
\\
\indent Moreover, the comparative analysis of these different methodologies shows the advantage of employing a \textit{before} and \textit{after} approach to satellite imagery analysis under a single input. By evaluating \textit{before} and \textit{after} images as coupled inputs, the model identifies changes that a single post-event imagery might not reveal, as shown by VGG16. The dual-image approach provides a better understanding of the extent and the specific locations of wildfire damage, offering a clear advantage over models that analyze images separately. 

\section{\uppercase{CONCLUSION}}
The methodology presented in this study enabled the efficient development of a large, bi-temporal labeled dataset, the California Wildfire GeoImaging Dataset (CWGID). By leveraging programmatic tools such as GEE and curated historical data, the need for manual intervention was reduced while generating a dataset with over 100,000 \textit{before} and \textit{after} image pairs. The CWGID proved to be a valuable resource for training several deep learning architectures, particularly the EfficientNet-B0 Early Fusion (EF) model. The bi-temporal nature of the dataset allowed this model to detect forest wildfires more accurately. The results of this study show that the CWGID can be used to  train DL architectures for detecting wildfires with high accuracy.
\\
\indent Future work will focus on using the dataset to train FCNs, such as U-Net, for pixel-wise wildfire detection, and extending our  methodology to other forms of environmental monitoring such as for illegal logging.
\bibliographystyle{apalike}
{\small
\bibliography{references}}

\begin{thebibliography}{}

\bibitem[Abujayyab et~al., 2023]{abujayyab2023wildfire}
Abujayyab, S.~K., Karas, I.~R., Hashempour, J., Emircan, E., Or{\c{c}}un, K., and Ahmet, G. (2023).
\newblock Wildfire detection from sentinel imagery using convolutional neural network (cnn).
\newblock In {\em The Proceedings of the International Conference on Smart City Applications}, pages 341--349. Springer.

\bibitem[Adegun et~al., 2023]{Adegun2023}
Adegun, A., Viriri, S., and Tapamo, J. (2023).
\newblock Review of deep learning methods for remote sensing satellite images classification: experimental survey and comparative analysis.
\newblock {\em Journal of Big Data}, 10:93.

\bibitem[Al-Dabbagh and Ilyas, 2023]{al-dabbagh2023uni}
Al-Dabbagh, A.~M. and Ilyas, M. (2023).
\newblock Uni-temporal sentinel-2 imagery for wildfire detection using deep learning semantic segmentation models.
\newblock {\em Geomatics, Natural Hazards and Risk}, 14(1).

\bibitem[Alice et~al., 2023]{alice2023automated}
Alice, K., Thillaivanan, A., Rao, G. R.~K., Rajalakshmi, S., Singh, K., and Rastogi, R. (2023).
\newblock Automated forest fire detection using atom search optimizer with deep transfer learning model.
\newblock In {\em 2023 2nd International Conference on Applied Artificial Intelligence and Computing (ICAAIC)}, pages 222--227. IEEE.

\bibitem[Alzubaidi et~al., 2021]{Alzubaidi2021ReviewOD}
Alzubaidi, L., Zhang, J., Humaidi, A.~J., Al-dujaili, A., Duan, Y., Al-Shamma, O., Santamar{\'i}a, J.~I., Fadhel, M.~A., Al-Amidie, M., and Farhan, L. (2021).
\newblock Review of deep learning: concepts, cnn architectures, challenges, applications, future directions.
\newblock {\em Journal of Big Data}, 8.

\bibitem[Arteaga et~al., 2020]{arteaga2020deep}
Arteaga, B., Diaz, M., and Jojoa, M. (2020).
\newblock Deep learning applied to forest fire detection.
\newblock In {\em 2020 IEEE International Symposium on Signal Processing and Information Technology (ISSPIT)}, pages 1--6. IEEE.

\bibitem[Berlinck and Batista, 2020]{berlinck2020good}
Berlinck, C.~N. and Batista, E.~K. (2020).
\newblock Good fire, bad fire: It depends on who burns.
\newblock {\em Flora}, 268:151610.

\bibitem[Brand and Manandhar, 2021]{brand2021semantic}
Brand, A. and Manandhar, A. (2021).
\newblock Semantic segmentation of burned areas in satellite images using a u-net-based convolutional neural network.
\newblock {\em The International Archives of the Photogrammetry, Remote Sensing and Spatial Information Sciences}, 43:47--53.

\bibitem[Bromley et~al., 1993]{bromley}
Bromley, J., Guyon, I., LeCun, Y., S\"{a}ckinger, E., and Shah, R. (1993).
\newblock Signature verification using a "siamese" time delay neural network.
\newblock In Cowan, J., Tesauro, G., and Alspector, J., editors, {\em Advances in Neural Information Processing Systems}, volume~6. Morgan-Kaufmann.

\bibitem[Davis and Shekaramiz, 2023]{davis}
Davis, M. and Shekaramiz, M. (2023).
\newblock Desert/forest fire detection using machine/deep learning techniques.
\newblock {\em Fire}, 6:1--20.

\bibitem[Drusch et~al., 2012]{DRUSCH201225}
Drusch, M., {Del Bello}, U., Carlier, S., Colin, O., Fernandez, V., Gascon, F., Hoersch, B., Isola, C., Laberinti, P., Martimort, P., Meygret, A., Spoto, F., Sy, O., Marchese, F., and Bargellini, P. (2012).
\newblock Sentinel-2: Esa's optical high-resolution mission for gmes operational services.
\newblock {\em Remote Sensing of Environment}, 120:25--36.
\newblock The Sentinel Missions - New Opportunities for Science.

\bibitem[Goodfellow et~al., 2016]{goodfellow2016deep}
Goodfellow, I., Bengio, Y., and Courville, A. (2016).
\newblock {\em Deep Learning}.
\newblock MIT Press, Cambridge, MA, USA.

\bibitem[Gorelick et~al., 2017]{gorelick2017google}
Gorelick, N., Hancher, M., Dixon, M., Ilyushchenko, S., Thau, D., and Moore, R. (2017).
\newblock Google earth engine: Planetary-scale geospatial analysis for everyone.
\newblock {\em Remote Sensing of Environment}.

\bibitem[Hu et~al., 2015]{hu2015transferring}
Hu, F., Xia, G.-S., Hu, J., and Zhang, L. (2015).
\newblock Transferring deep convolutional neural networks for the scene classification of high-resolution remote sensing imagery.
\newblock {\em Remote Sensing}, 7(11):14680--14707.

\bibitem[Hutto, 2008]{hutto2008ecological}
Hutto, R.~L. (2008).
\newblock The ecological importance of severe wildfires: some like it hot.
\newblock {\em Ecological Applications}, 18(8):1827--1834.

\bibitem[{International Union for Conservation of Nature}, 2021]{IUCN2021}
{International Union for Conservation of Nature} (2021).
\newblock Forests and climate change.
\newblock \url{https://www.iucn.org/resources/issues-brief/forests-and-climate-change}.
\newblock Accessed: 2024-02-28.

\bibitem[Jiang et~al., 2022]{rs14071552}
Jiang, H., Peng, M., Zhong, Y., Xie, H., Hao, Z., Lin, J., Ma, X., and Hu, X. (2022).
\newblock A survey on deep learning-based change detection from high-resolution remote sensing images.
\newblock {\em Remote Sensing}, 14(7).

\bibitem[Khan et~al., 2022]{deepfire}
Khan, A., Hassan, B., Khan, S., Ahmed, R., and Abuassba, A. (2022).
\newblock Deepfire: A novel dataset and deep transfer learning benchmark for forest fire detection.
\newblock {\em Mobile Information Systems}, 2022.

\bibitem[Kingma and Ba, 2014]{kingma2014adam}
Kingma, D.~P. and Ba, J. (2014).
\newblock Adam: A method for stochastic optimization.
\newblock {\em arXiv preprint arXiv:1412.6980}.

\bibitem[Knopp et~al., 2020]{knopp2020deep}
Knopp, L., Wieland, M., R{\"a}ttich, M., and Martinis, S. (2020).
\newblock A deep learning approach for burned area segmentation with sentinel-2 data.
\newblock {\em Remote Sensing}, 12(15):2422.

\bibitem[Lecun and Bengio, 1995]{lecun}
Lecun, Y. and Bengio, Y. (1995).
\newblock Convolutional networks for images, speech, and time-series.

\bibitem[Li and Banerjee, 2021]{li2021spatial}
Li, S. and Banerjee, T. (2021).
\newblock Spatial and temporal pattern of wildfires in california from 2000 to 2019.
\newblock {\em Scientific reports}, 11(1):8779.

\bibitem[Marmanis et~al., 2016]{marmanis2016deep}
Marmanis, D., Datcu, M., Esch, T., and Stilla, U. (2016).
\newblock Deep learning earth observation classification using imagenet pretrained networks.
\newblock {\em IEEE Geoscience and Remote Sensing Letters}, 13(1):105--109.

\bibitem[Massey et~al., 2023]{Massey2023}
Massey, R., Berner, L.~T., Foster, A.~C., Goetz, S.~J., and Vepakomma, U. (2023).
\newblock {\em Remote Sensing Tools for Monitoring Forests and Tracking Their Dynamics}, pages 637--655.
\newblock Springer International Publishing, Cham.

\bibitem[Parelius, 2023]{eleo}
Parelius, E.~J. (2023).
\newblock A review of deep-learning methods for change detection in multispectral remote sensing images.
\newblock {\em Remote Sensing}, 15(8).

\bibitem[Perez and Wang, 2017]{perez2017effectiveness}
Perez, L. and Wang, J. (2017).
\newblock The effectiveness of data augmentation in image classification using deep learning.
\newblock {\em arXiv preprint arXiv:1712.04621}.

\bibitem[Pinto et~al., 2021]{pinto2021practical}
Pinto, M.~M., Trigo, R.~M., Trigo, I.~F., and DaCamara, C.~C. (2021).
\newblock A practical method for high-resolution burned area monitoring using sentinel-2 and viirs.
\newblock {\em Remote Sensing}, 13(9):1608.

\bibitem[Ronneberger et~al., 2015]{DBLP:RonnebergerFB15}
Ronneberger, O., Fischer, P., and Brox, T. (2015).
\newblock U-net: Convolutional networks for biomedical image segmentation.
\newblock {\em CoRR}, abs/1505.04597.

\bibitem[Sawyer et~al., 2009]{mcv2}
Sawyer, J., Keeler-Wolf, T., and Evens, J. (2009).
\newblock {\em Manual of California Vegetation}.
\newblock California Native Plant Society, 2 edition.

\bibitem[Seydi et~al., 2022]{SEYDI2022108999}
Seydi, S.~T., Hasanlou, M., and Chanussot, J. (2022).
\newblock Burnt-net: Wildfire burned area mapping with single post-fire sentinel-2 data and deep learning morphological neural network.
\newblock {\em Ecological Indicators}, 140:108999.

\bibitem[Shorten and Khoshgoftaar, 2019]{shorten19}
Shorten, C. and Khoshgoftaar, T.~M. (2019).
\newblock A survey on image data augmentation for deep learning.
\newblock {\em Journal of Big Data}, 6(1):60.

\bibitem[Simonyan and Zisserman, 2015]{Simonyan15}
Simonyan, K. and Zisserman, A. (2015).
\newblock Very deep convolutional networks for large-scale image recognition.
\newblock In {\em International Conference on Learning Representations}.

\bibitem[Sun, 2022]{sun2022analyzing}
Sun, C. (2022).
\newblock Analyzing multispectral satellite imagery of south american wildfires using deep learning.
\newblock In {\em 2022 International Conference on Applied Artificial Intelligence (ICAPAI)}, pages 1--6. IEEE.

\bibitem[Tan and Le, 2019]{tan2019efficientnet}
Tan, M. and Le, Q.~V. (2019).
\newblock Efficientnet: Rethinking model scaling for convolutional neural networks.
\newblock In {\em Proceedings of the 36th International Conference on Machine Learning (ICML)}, pages 6105--6114.

\bibitem[Vilar~del Hoyo et~al., 2007]{vilar2007analisis}
Vilar~del Hoyo, L. et~al. (2007).
\newblock An{\'a}lisis comparativo de diferentes m{\'e}todos para la obtenci{\'o}n de modelos de riesgo humano de incendios forestales: Madrid.

\bibitem[Xiang et~al., 2023]{Hunan}
Xiang, J., Xing, Y., Wei, W., Yan, E., Jiang, J., and Mo, D. (2023).
\newblock Dynamic detection of forest change in hunan province based on sentinel-2 images and deep learning.
\newblock {\em Remote Sensing}, 15(3).

\bibitem[Xie et~al., 2016]{xie2016deep3d}
Xie, J., Girshick, R., and Farhadi, A. (2016).
\newblock Deep3d: Fully automatic 2d-to-3d video conversion with deep convolutional neural networks.
\newblock In {\em European Conference on Computer Vision}, pages 842--857. Springer.

\bibitem[Zhang et~al., 2024]{zhang2024assessing}
Zhang, P., Hu, X., Ban, Y., Nascetti, A., and Gong, M. (2024).
\newblock Assessing sentinel-2, sentinel-1, and alos-2 palsar-2 data for large-scale wildfire-burned area mapping: Insights from the 2017--2019 canada wildfires.
\newblock {\em Remote Sensing}, 16(3):556.

\bibitem[Zhang et~al., 2021]{zhang2021deep}
Zhang, Q., Ge, L., Zhang, R., Metternicht, G.~I., Du, Z., Kuang, J., and Xu, M. (2021).
\newblock Deep-learning-based burned area mapping using the synergy of sentinel-1\&2 data.
\newblock {\em Remote Sensing of Environment}, 264:112575.

\bibitem[Zhu et~al., 2017]{8113128}
Zhu, X.~X., Tuia, D., Mou, L., Xia, G.-S., Zhang, L., Xu, F., and Fraundorfer, F. (2017).
\newblock Deep learning in remote sensing: A comprehensive review and list of resources.
\newblock {\em IEEE Geoscience and Remote Sensing Magazine}, 5(4):8--36.

\end{thebibliography}

\end{document}